\documentclass[runningheads]{llncs}
\usepackage{graphicx}
\usepackage{cite} 
\usepackage{times}
\usepackage{epsfig}
\usepackage{amsmath}
\usepackage{amssymb}
\usepackage{mwe}
\usepackage{acro}
\usepackage{amssymb}
\usepackage{xcolor,colortbl}
\usepackage{tabularx}
\usepackage{relsize}
\usepackage{pifont}
\usepackage{booktabs} 
\usepackage{multirow}
\usepackage{multicol}
\usepackage{adjustbox}
\usepackage{float}

\usepackage{cite}
\usepackage{amsmath,amssymb,amsfonts}
\usepackage{graphicx}
\usepackage{textcomp}

\usepackage{algorithmicx}               
\usepackage{algorithm} 
\usepackage{algpseudocode}  

\usepackage{array}
\usepackage{booktabs}
\usepackage{colortbl}
\definecolor{dark-gray}{gray}{0.7} 
\usepackage{multirow}

\usepackage{float}
\usepackage{cuted}
\usepackage{graphicx, caption}
\usepackage[colorlinks,
            linkcolor=red,  
            anchorcolor=blue, 
            citecolor=green,   
            ]{hyperref}

\DeclareAcronym{ROI}{
short=ROI,
long=region of interest,
}

\DeclareAcronym{IOU}{
short=IOU,
long=intersection over union,
}

\DeclareAcronym{cIOU}{
short=cIOU,
long=circle intersection over union,
}

\DeclareAcronym{DoF}{
short=DoF,
long=degrees of freedom,
}

\begin{document}
\title{CaCL: class-aware codebook learning for weakly supervised segmentation on diffuse image patterns}
%
%
\author{Ruining Deng\inst{1} \and
Quan Liu \inst{1} \and
Shunxing Bao \inst{1} \and
Aadarsh Jha \inst{1} \and
Catie Chang \inst{1} \and
Bryan A. Millis \inst{2} \and
Matthew J.Tyska \inst{2} \and
Yuankai Huo\inst{1}}


%
\institute{
Vanderbilt University, Nashville TN 37215, USA \and
Vanderbilt University Medical Center, Nashville TN 37215, USA
}

%
\maketitle              
\begin{abstract}
Weakly supervised learning has been rapidly advanced in biomedical image analysis to achieve pixel-wise labels (segmentation) from image-wise annotations (classification), as biomedical images naturally contain image-wise labels in many scenarios. The current weakly supervised learning algorithms from the computer vision community are largely designed for focal objects (e.g., dogs and cats). However, such algorithms are not optimized for diffuse patterns in biomedical imaging (e.g., stains and fluorescence in microscopy imaging). In this paper, we propose a novel class-aware codebook learning (CaCL) algorithm to perform weakly supervised learning for diffuse image patterns. Specifically, the CaCL algorithm is deployed to segment protein expressed brush border regions from histological images of human duodenum. Our contribution is three-fold: (1) we approach the weakly supervised segmentation from a novel codebook learning perspective; (2) the CaCL algorithm segments diffuse image patterns rather than focal objects; and (3) the proposed algorithm is implemented in a multi-task framework based on Vector Quantised-Variational AutoEncoder (VQ-VAE) via joint image reconstruction, classification, feature embedding, and segmentation. The experimental results show that our method achieved superior performance compared with baseline weakly supervised algorithms. The code is available at  \url{https://github.com/ddrrnn123/CaCL}
\end{abstract}
\section{Introduction}
Mapping the location of 19,628 human protein‐coding genes plays a critical role as a “census” of proteins, which further increases our knowledge of human biology and enables new insights into principles of life. For instance, the Human Protein Atlas (HPA) project\footnote{https://www.proteinatlas.org} has applied $>$25,000 antibodies to characterize the tissue-level spatial expression by collecting 10 million immunohistochemistry (IHC) images. The IHC images indicate the location and distribution of protein expression. For example, understanding the area ratio between IHC stained regions and cell body regions at the brush border of the human duodenum reveals the specificity of gene expressions. 

The color deconvolution algorithm~\cite{ruifrok2001quantification} is regarded as the de facto standard approach to segment IHC stained histopathology images. However, the manual tuning of IHC staining parameters (e.g., segmentation threshold) to deal with heterogeneous image qualities and attributes is labor-intensive. Moreover, color deconvolution cannot understand the semantic information of a figure.

\begin{figure}[t]
\begin{center}
\includegraphics[width=4 in]{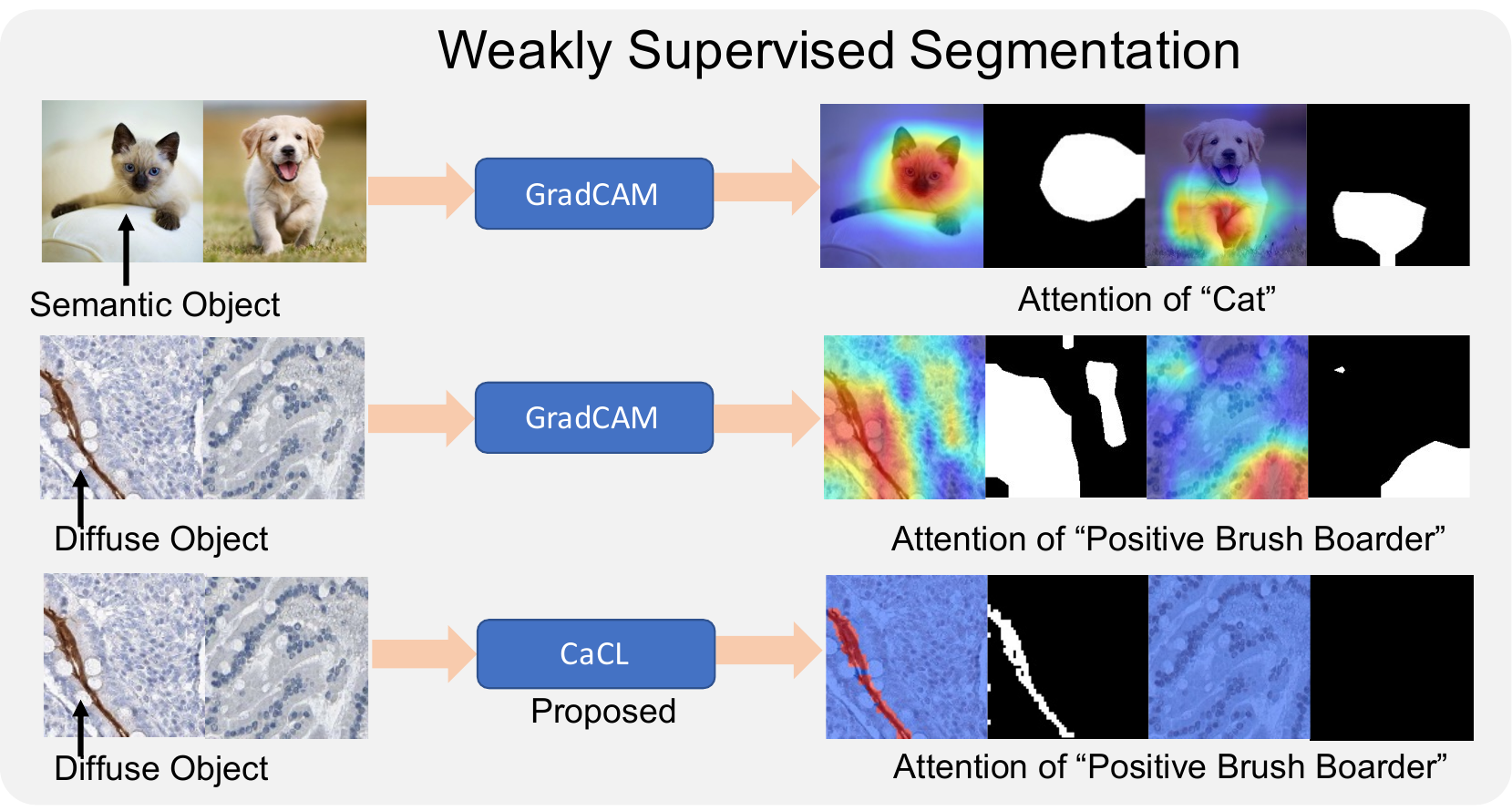}
\end{center}
   \caption{\textbf{The performances of object segmentation.} This figure shows the performances of object segmentation using different attention-based weakly supervised learning methods. The former method, GradCAM, is designed for focal objects rather than diffuse objects. Our proposed method, CaCL, can obtain better results on diffuse objects.}
\label{fig1}
\end{figure}

Recent weakly supervised learning techniques have played a critical role in image segmentation with the benefits of only needing image-wise annotation~\cite{wang2019weakly}. Zhou et al.,~\cite{zhou2016learning} proposed Class Activation Mapping (CAM) for Convolutional Neural Network(CNN) with a global average pool to allow CNNs to visualize object localization. Then, Selvaraju et al.,~\cite{selvaraju2017grad} developed Gradient-weighted Class Activation Mapping (GradCAM) and Guided GradCAM (G-GradCAM) for better visual explanations with localization information. Later on, Fong et al.,~\cite{fong2017interpretable,fong2019understanding} introduced a framework for learning meta-predictors. However, the current weakly supervised learning algorithms from the computer vision community are mostly designed for focal objects and may display attention with any image, which are not optimized for diffuse patterns in biomedical imaging~\cite{chan2021comprehensive}(Fig. \ref{fig1}). 

Meanwhile, there have been several weakly supervised learning approaches in histology~\cite{rony2019deep}. Belharbi et al.,~\cite{belharbi2021deep} proposed an active learning framework to jointly perform supervised image-level classification and active learning for segmentation. Xu et al., ~\cite{xu2019camel} proposed a weakly supervised learning framework for histopathology image segmentation, using multiple instance learning (MIL)-based label enrichment and fully supervised training with image-level labels. These methods achieved superior performances. With the development of deep learning technology, more unsupervised segmentation models were proposed for medical image analysis~\cite{baumgartner2018visual,baur2018deep,you2019unsupervised,pawlowski2018unsupervised}. However, most of the attention-based methods only obtained attention maps for partial classification tasks rather than segmentation tasks. Herein, we provided a weakly supervised learning model to achieve robust segmentation images directly from attention maps.

A new generative model, Vector Quantised-Variational AutoEncoder (VQ-VAE)~\cite{oord2017neural},  was proposed to encode an image from an infinite continuous feature space to a finite discrete feature space using a codebook with a fixed number of codes. Inspired by VQ-VAE, we propose a novel class-aware codebook learning (CaCL) algorithm to segment diffuse patterns in medical imaging. The central idea is to split the original codebook into two separate codebooks. One codebook encodes the discriminative class patterns (codebook $C$), while the other encodes the common image patterns between two groups of images (codebook $S$). Then, the pixels that used in the codebook $S$ + $C$ during the encoding process are used as an “attention” to perform weakly supervised segmentation. Briefly, the innovations of the proposed approach is in three-fold: (1) We approach the weakly supervised segmentation from a novel codebook learning perspective; (2) We introduce the CaCL algorithm to segment diffuse image patterns rather than focal objects; (3) The proposed algorithm is implemented in a multi-task framework based on Vector Quantised-Variational AutoEncoder (VQ-VAE) via joint image reconstruction, classification, feature embedding, and segmentation.

\section{Methods}
\begin{figure}[t]
\begin{center}
\includegraphics[width=4.5 in]{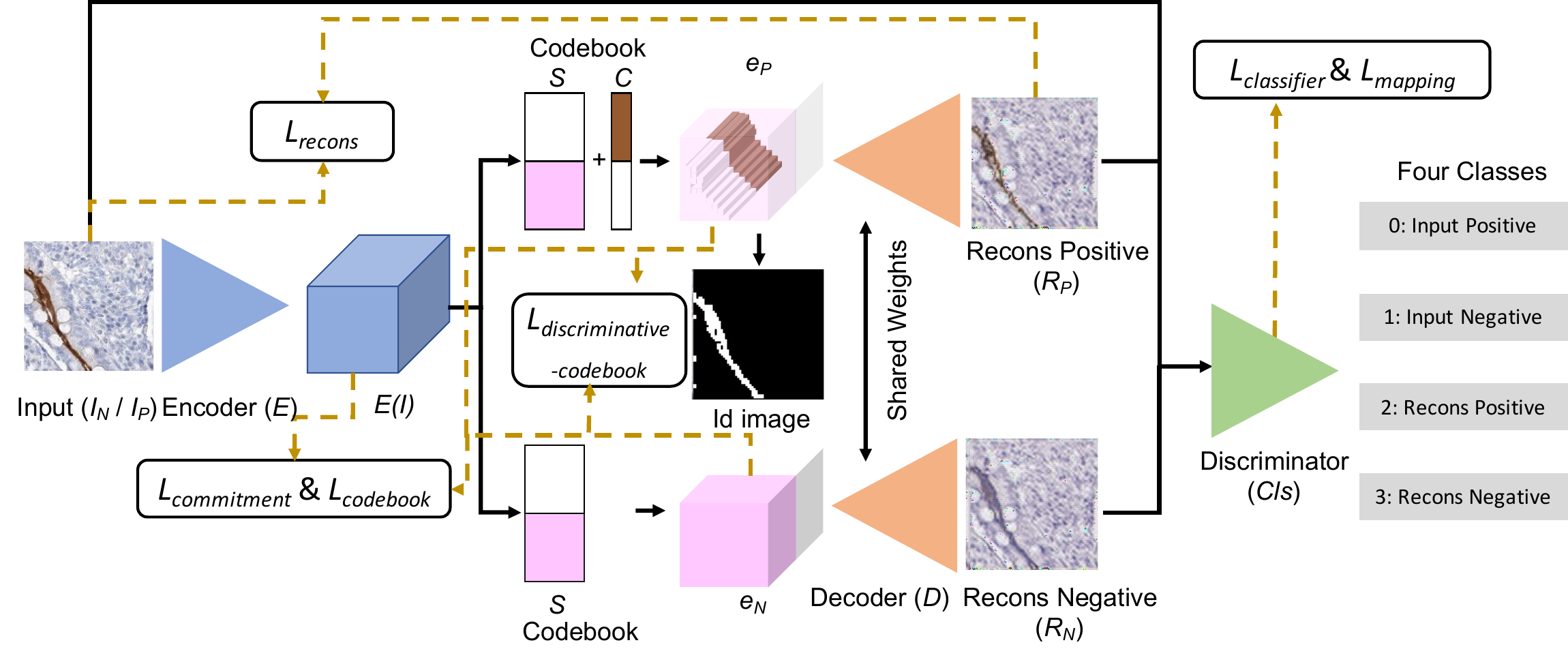}
\end{center}
\caption{\textbf{The backbone of our method.} Our method includes CaCL embedding, GAN based reconstruction and classification, and weakly supervised learning segmentation.}
\label{fig2}
\end{figure}

The entire framework of the proposed CaCL method is presented in Fig. \ref{fig2} . The CaCL algorithm consists of three sections: (1) a class-aware codebook for feature embedding; (2) generative adversarial image reconstruction and classification; and (3) weakly supervised segmentation from diffuse patterns.

\subsection{Class-aware Codebook Based Feature Encoding} 
In this study, we design a class-aware codebook inspired by VQ-VAE2~\cite{razavi2019generating}. With the VQ-VAE framework, three steps are used to process an input image. First, the encoder $E$ is used to convert a RGB image into a feature map. Second, the feature map is coded by the codebook from an infinite solution space to a fixed number of codes for each pixel. For example, if the codebook contains 32 codes, each pixel can only be one of the 32 types of features. Last, the coded feature maps were decoded to the input image resolution as a encoder-decoder design.

\begin{figure}[t]
\begin{center}
\includegraphics[width=2 in]{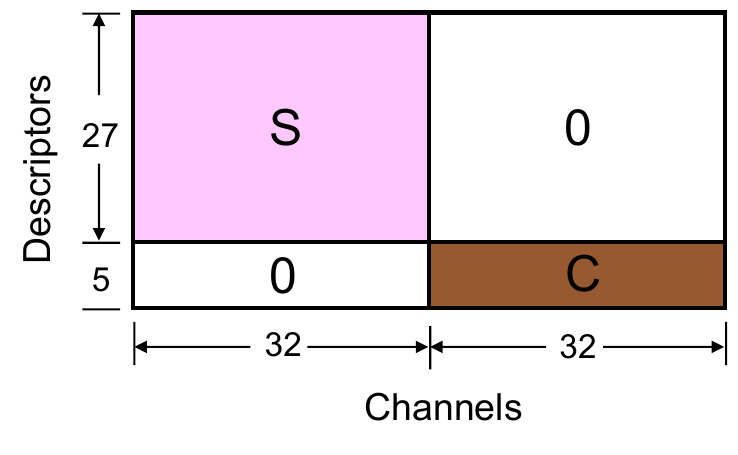}
\end{center}

\caption{\textbf{The design of the class-aware codebook.} This figure shows the design of the class-aware codebook. One encodes class discriminative features (codebook $C$), while another encodes the shared features among two classes (codebook $S$).}
\label{fig5}
\end{figure}

As opposed to VQ-VAE, which only used one codebook to encode all inputs, we propose to use two codebooks in CaCL. One encodes class discriminative features (codebook $C$), while another encodes the shared features among two classes (codebook $S$), as shown in Fig. \ref{fig5}. In this study, the images with positive protein expression patterns (dark brown at the brush broader) are defined as $I_P$, while the images without protein expression patterns are defined as $I_N$. Then, for each input image, we will first retrieve one raw feature map from $E$. Second, two coded feature maps will be obtained by using codebook $S$ only and codebook $S+C$, respectively. Two images will be reconstructed using the same decoder $D$. One image only has common diffuse patterns across positive and negative images (Recons Negative $R_N$ in Fig. \ref{fig2}), while another image contains both common diffuse patterns and class discriminative patterns (Recons Positive $R_P$ in Fig. \ref{fig2}).

\subsection{Loss Definition}
\label{ssec:subhead}

\textbf{Commitment loss and codebook loss:} 
 Herein, we implement the commitment loss and codebook loss in VQ-VAE2 that retains the reconstruction features close to the chosen codebook vectors.

\begin{equation}\label{eq:commitment}
\begin{aligned}
\noindent \mathcal{L}_{commitment}&(I,D(e))= ||sg[e] - E(I)||_{2}^{2} 
\end{aligned}
\end{equation}

\begin{equation}\label{eq:codebook}
\begin{aligned}
\noindent \mathcal{L}_{codebook}&(I,D(e))= ||sg[E(I)] - e||_{2}^{2}
\end{aligned}
\end{equation}

\noindent where $e$ is the coded feature map for the input $I$, $E$ is the encoder function, and $D$ is the decoder function. The operator $sg$ refers to a non-gradient operation that stops the gradients from flowing into its argument. It uses the exponential moving average updates for the codebook with a decay parameter.

\textbf{Reconstructive loss:} 
The reconstructive loss is applied to supervise the quality of reconstruction images $R_P$ and $R_N$. Each input image $I$ will go through the combined codebook $C$ and $S$ and the single codebook $S$, which obtain both $R_P$ and $R_N$. $R_P$ is calculated through the mean-square-error as the reconstruction loss with $I_P$, and $R_N$ is compared with images $I_N$, respectively.

\begin{equation}\label{eq:recons}
\begin{aligned}
    \noindent \mathcal{L}_{recons}&(I,R_P,R_N) = (1-M)||I-R_P||_{2}^{2} + M||I-R_N||_{2}^{2}\\
    \noindent&\text{ Where } M = 
    \begin{cases}
    1, \quad I = I_N\\
    0, \quad I = I_P
    \end{cases}
\end{aligned}
\end{equation}

\textbf{Discriminative-codebook loss:} 
To encourage the model to use codebook $C$, we introduce a new discriminative-codebook loss to calculate the mean-square-error of the quantized feature maps $e_N$ and $e_P$ in the non-zero channels from codebook $C$. Briefly, if the image is negative, we force the feature maps to be identical from two code books. If the image is positive, we force the feature maps to be different from two code books by using $\mathcal{L}_{discriminative-codebook}$.

\begin{equation}\label{eq:discriminative-codebook}
\begin{aligned}
\noindent \mathcal{L}_{discriminative-codebook}&(I,e_N,e_P) = K||e_N-e_P||_{2}^{2}\\
\noindent&\text{ Where } K = 
\begin{cases}
1, \quad I = I_N\\
-1, \quad I = I_P
\end{cases}
\end{aligned}
\end{equation}

\textbf{Hybrid discriminator loss}:
The hybrid discriminator loss performs both: (1) real/fake; and (2) positive/negative classification tasks on reconstructed images. The implementation of the discriminator and the generator are followed by a generative adversarial network (GAN) design~\cite{zhu2017unpaired}. We create two image pools $P_{data}$ to separately store all fake positive and fake negative images to train the discriminator. We use one resnet18~\cite{he2016deep}, named as $Cls$, as the discriminator (classifier). 

\begin{equation}\label{eq:classifier}
\begin{aligned}
\mathcal{L}_{classifier}&(I_P,I_N,R_P,R_N) =  T_{R_{P}}log(Cls(X\sim P_{data(R_P)}))\\
& + T_{R_{N}}log(Cls(X\sim P_{data(R_N)})) \\
& + T_{I_{P}}log(Cls(I_P)) + T_{I_{N}}log(Cls(I_N))
\end{aligned}
\end{equation}

\begin{equation}\label{eq:mapping}
\begin{aligned}
\mathcal{L}_{mapping}(R_P,R_N) = T_{I_{P}}log(Cls(R_P)) +  T_{I_{N}}log(Cls(R_N))
\end{aligned}
\end{equation}

\noindent where $T_{I_{P}}$, $T_{I_{N}}$, $T_{R_{P}}$, $T_{R_{N}}$ are the targets of $I_P$, $I_N$, $R_P$, $R_N$, respectively.   

The aforementioned loss functions are aggregated into $\mathcal{L}_{combine}$ with weights $\lambda$. Since the discriminators typically converge faster than generators, we perform back-propagation at different frequencies. During the training, the classification loss ($\mathcal{L}_{classifier}$) is updated in every ten epochs, while $\mathcal{L}_{combine}$ is updated in each epoch.

\begin{equation}\label{eq:combined}
\begin{aligned}
\mathcal{L}_{combine}& = \lambda_{mapping}\mathcal{L}_{mapping} + \lambda_{commitment}\mathcal{L}_{commitment} + \lambda_{recons}\mathcal{L}_{recons}\\
& + \lambda_{discriminative-codebook}\mathcal{L}_{discriminative-codebook} \\
\end{aligned}
\end{equation}

\subsection{Training Strategy}
The class consistency is normalized by computing commitment loss and reconstructive loss. For all positive images ($I_p$), only reconstructed positive images ($R_p$) are calculated in reconstructive loss. The raw features $E(I_p)$ from positive input images ($I_p$) are computed in commitment loss with positive coded features ($e_p$). The same principles are implemented for all negative inputs ($I_n$).

To train the codebooks, all the vectors in both codebook $S$ and $C$ are updated after quantizing the coded positive features ($e_p$). In contrast, only the vectors in codebook $S$ are updated after quantizing the coded negative features ($e_n$). Meanwhile, we use mean-square-error to reduce the difference between encoded features ($e_p$ and $e_n$) from the negative inputs ($I_n$), while simultaneously amplifying the distinctions between $e_p$ and $e_n$ from the positive inputs in the discriminative-codebook loss $\mathcal{L}_{discriminative-codebook}$. Such a process guides the codebook $S$ and codebook $C$ to assemble distinctive features in different classes, independently.

Next, a classifier is used to identify four types of images, which are Input Positive ($I_p$), Input Negative ($I_n$), Reconstructive Positive ($R_p$), and Reconstructive Negative($R_n$). Meanwhile, we employ a discriminator to reconcile the differences between the input images ($I_p$, $I_n$) and the reconstructive images ($R_p$, $R_n$). Ideally, only $R_p$ from $I_p$ contain the positive patterns from the codebook $C$. 

\subsection{Weakly Supervised Learning Segmentation}
Ideally, after training the model, only pixels using class-specific codebook $C$ should contribute to the differences between the two classes. Therefore, we simply mark those pixels as 1, and mark the remaining pixels as 0. The outcome mask is used as our weakly supervised segmentation results.

\section{Data and Experiments}

This research study was conducted retrospectively using human subject data made available in open access by the Human Protein Atlas (https://www.proteinatlas.org). Ethical approval was not required as confirmed by the license attached with the open access data. 42 high resolution duodenum histological micro-array images were obtained from the Human Protein Atlas. 27 images contained high brush border protein expression, while the remaining ones did not. The protein expression is specified as the IHC staining pattern (dark brown). Patches in an 8$\times$8 grid without overlapping from each high-resolution image were extracted. Due to the GPU memory limitation, we downsample these patches with 375$\times$375 pixels to image patches with 128$\times$128 pixels. We randomly selected 1480 patches for training, 200 patches for validation, and 200 for testing. Half of the testing images were from the brush border area to evaluate the performance of our method.

The design of the class-aware codebook is in Fig. \ref{fig5}. The number of descriptors in codebook $S$ is 27, while the number of descriptors in codebook $C$ is 5. Each descriptor has 64 channels, where 32 non-overlapping channels are from each codebook $S$ and $C$, respectively. The remaining locations of the codebook are set to 0. The decay in each codebook update is 0.98. For all the experiments, we use the Adam solver for optimization with a batch size of one. The learning rate of the classification loss is 0.0001, while the learning rate of the combined loss is 0.0003. The size of the image pool is 64. The weights $\lambda$ of commitment loss, reconstructive loss, discriminative-codebook loss, and discriminator loss are empirically set to 0.25,100,50, and 1, respectively. These parameters were determined by fine-tuning process to obtain superior performances in both segmentation metrics and reconstructive visualizations.

The color deconvolution was employed as the current standard IHC stain segmentation method. CAM and GradCAM were utilized as the benchmarks of attention based weakly supervised learning. All experiments were completed on the same workstation, with NVIDIA Quadro P5000 GPU.

\section{Results}

\begin{figure}[t]
\begin{center}
\includegraphics[width=3 in]{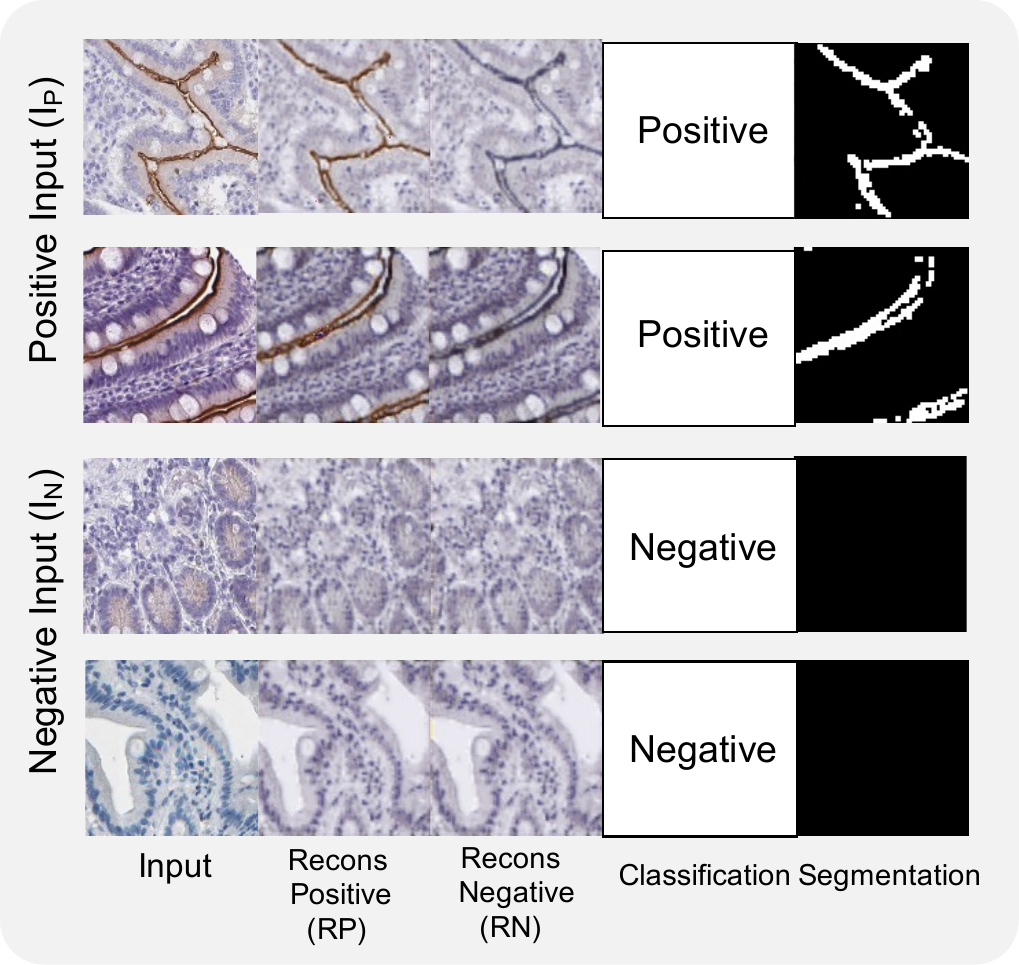}
\end{center}
\caption{\textbf{CaCL at the testing stage.} This figure shows the example outcomes from the proposed CaCL framework, which include reconstructed images, classification results and segmentation results.}
\label{fig3}
\end{figure}

\begin{figure}[t]
\begin{center}
\includegraphics[width=4.5 in]{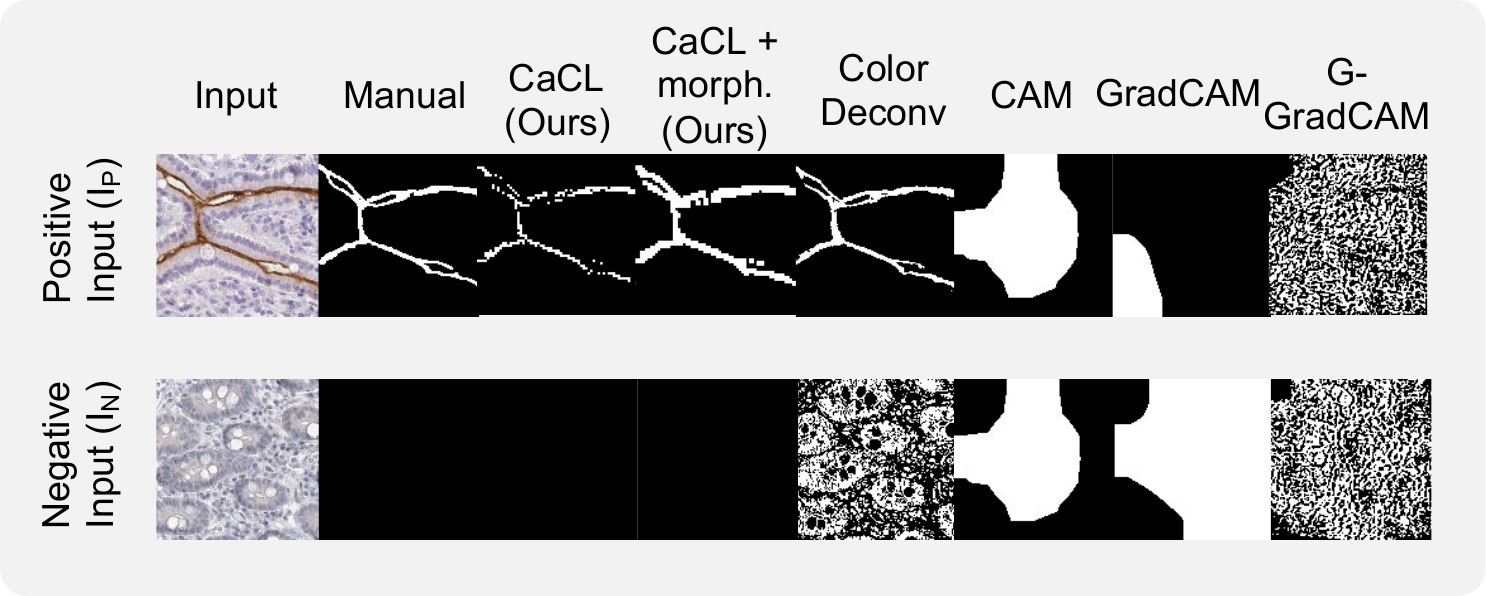}
\end{center}
\caption{\textbf{Pixel-wise attention segmentation.} This figure shows the results of brush border segmentation using pixel-wise attention from different weakly supervised learning methods.}
\label{fig4}
\end{figure}

In Fig.\ref{fig3}, the example input $I_P$ and $I_N$ images, and the corresponding reconstructed $R_P$ and $R_N$, are presented. Fig. \ref{fig4} shows the qualitative weakly segmentation results, while Table \ref{table1} presents the quantitative results. The Dice Similarity Coefficient (DSC), Positive Predictive Value (Precision), Sensitivity (Recall), and Binary Cross-Entropy (BCE) are used as evaluation metrics. For each $I_N$ image, if all the pixels inside the predicted segmentation masks are 0, then DSC, Precision, and Recall are computed as 1. Otherwise, those metrics are 0, according to~\cite{chicco2020advantages}. All the results of baseline models in Table \ref{table1} are the best performances by iterating all the intensity values as thresholds. A simple morphological dilation operation with radius 1 is also tested in Table \ref{table1}. As a result, our method achieved the best quantitative performance.

\begin{table}[t]
\centering
\caption{Segmentation results.}
\begin{tabular}{cccccccccccccccccc}
\toprule
\bfseries Method& \bfseries Dice& \bfseries Recall& \bfseries Precision & \bfseries BCE \\
\midrule
Color deconv.~\cite{ruifrok2001quantification}&0.347&0.400&0.363&7.066\\
CAM~\cite{zhou2016learning}&0.065&0.038&0.260&15.813\\
GradCAM~\cite{selvaraju2017grad}&0.061&0.035&0.298&19.586\\
G-GradCAM~\cite{selvaraju2017grad}&0.030&0.018&0.099&14.273\\
CaCL (Ours) &0.623&\textbf{0.787}&0.574&\textbf{1.079}\\
CaCL+morph. (Ours)&\textbf{0.703}&0.712&\textbf{0.723}&1.258\\
\bottomrule
\end{tabular}
\label{table1}
\end{table}

\section{Discussion}
In this study, we presented a new weakly supervised learning method with a class-aware codebook. The proposed CaCL approach achieved diffuse pattern segmentation without pixel-wise annotation. Our proposed method combines with the classification task and the segmentation task as a whole with ``pixel-wise attention" from image-wise weak labels, while previous CAM based attention is more coerce.

The codebook-based reconstruction uses the region-level features from pixel-wise feature maps, which inhibit positive features. The purpose of the dilation enhancement is to decrease this impact from neighbor pixels, which achieve better segmentation results in Table \ref{table1}.

The goal of our method is to achieve both focal pattern segmentation and controlling the expression of positive patterns with the realistic reconstructive images by class-aware codebooks. In our experiment, simply using the standard reconstruction loss without the discriminator loss $\mathcal{L}_{mapping}$ generates numerous unreasonable noise pixels as fake patterns on the reconstructive images, which can cheat in the classifier $\mathcal{L}_{classifier}$ and fail to control the pattern expression. In Fig. \ref{fig4}, our design can receive segmentation results while achieving consistent expression control in IHC stained histopathology images. 

At current stage, there are still major limitations. One obvious limitation is the size of our dataset. The number of the training and testing images is limited due to the limitation of resources and extensive labor costs, as well as time needed to achieve pixel-wise manual annotations. More training data would lead to better segmentation performance. In the future, one promising improvement of the proposed method would be to extend the current binary classification and segmentation approach to multi-class scenarios.






%
%
\bibliographystyle{splncs04}
\bibliography{main}
%




\end{document}